\documentclass[10pt,twocolumn,letterpaper]{article}

\usepackage{cvpr}              % To produce the CAMERA-READY version
\usepackage{algorithm}
\usepackage{algpseudocode}   % not 'algorithmic'
\usepackage{booktabs}
\usepackage[table]{xcolor}
\usepackage{makecell}
\usepackage{placeins}

\algrenewcommand\alglinenumber[1]{\scriptsize #1:}
\definecolor{cvprblue}{rgb}{0.21,0.49,0.74}
\usepackage[pagebackref,breaklinks,colorlinks,allcolors=cvprblue]{hyperref}
\algrenewcommand\algorithmicrequire{\textbf{Inputs:}}
\algrenewcommand\algorithmicensure{\textbf{Hparams:}}

\title{EBT-Policy: Energy Unlocks Emergent Physical Reasoning Capabilities}

\author{
Travis Davies\\
ZhiCheng AI\\
\and
Yiqi Huang \\
ZhiCheng AI \\
\and 
Alexi Gladstone\\
UIUC\\
\and 
Yunxin Liu \\
Tsinghua University \\
\and 
Xiang Chen \\
Peking University \\
\and
Heng Ji \\
UIUC
\and
Huxian Liu \\
ZhiCheng AI \\
\and
Luhui Hu \\
ZhiCheng AI
}

\begin{document}
\maketitle
\begin{abstract}
% The ABSTRACT is to be in fully justified italicized text, at the top of the left-hand column, below the author and affiliation information.
% Use the word ``Abstract'' as the title, in 12-point Times, boldface type, centered relative to the column, initially capitalized.
% The abstract is to be in 10-point, single-spaced type.
% Leave two blank lines after the Abstract, then begin the main text.
% Look at previous \confName abstracts to get a feel for style and length.

% TODO mention best results, e.g. more than 50x more efficient than DP while using less than half the training epochs; the first BC approach we've ever seen being able to recover from missed placements without ever seeing that in the training data or being taught to do that, same flow as intro

% In this work we introduce a new approach towards implicit policies which 

% Implicit Policies parametrized by an Energy-Based Model (EBMs) offer high promise in handling uncertainty, are capable of dynamic reasoning, and generalize well under covariate shift~\cite{ibc}. 
\vspace{-5pt}
Implicit policies parameterized by generative models, such as Diffusion Policy~\cite{dp}, have become the standard for policy learning and Vision–Language–Action (VLA) models~\cite{openvla} in robotics. However, these approaches often suffer from high computational cost, exposure bias, and unstable inference dynamics, which lead to divergence under distribution shifts. Energy-Based Models (EBMs)~\cite{ebm-tutorial} address these issues by learning energy landscapes end-to-end and modeling equilibrium dynamics, offering improved robustness and reduced exposure bias. Yet, policies parameterized by EBMs have historically struggled to scale effectively. Recent work on Energy-Based Transformers (EBTs)~\cite{ebt} demonstrates the scalability of EBMs to high-dimensional spaces, but their potential for solving core challenges in physically embodied models remains underexplored. We introduce a new energy-based architecture, EBT-Policy, that solves core issues in robotic and real-world settings. Across simulated and real-world tasks, EBT-Policy consistently outperforms diffusion-based policies, while requiring less training and inference computation. Remarkably, on some tasks it converges within just two inference steps, a 50$\times$ reduction compared to Diffusion Policy’s 100. Moreover, EBT-Policy exhibits emergent capabilities not seen in prior models, such as zero-shot recovery from failed action sequences using only behavior cloning and without explicit retry training. By leveraging its scalar energy for uncertainty-aware inference and dynamic compute allocation, EBT-Policy offers a promising path toward robust, generalizable robot behavior under distribution shifts.\footnote{This is part of a broader effort on world models for robot manipulation; future updates and hyperparameter tuning are expected to further improve performance and results.}
\end{abstract}

\begin{figure}[t]
    \centering
    \includegraphics[width=0.9\columnwidth]{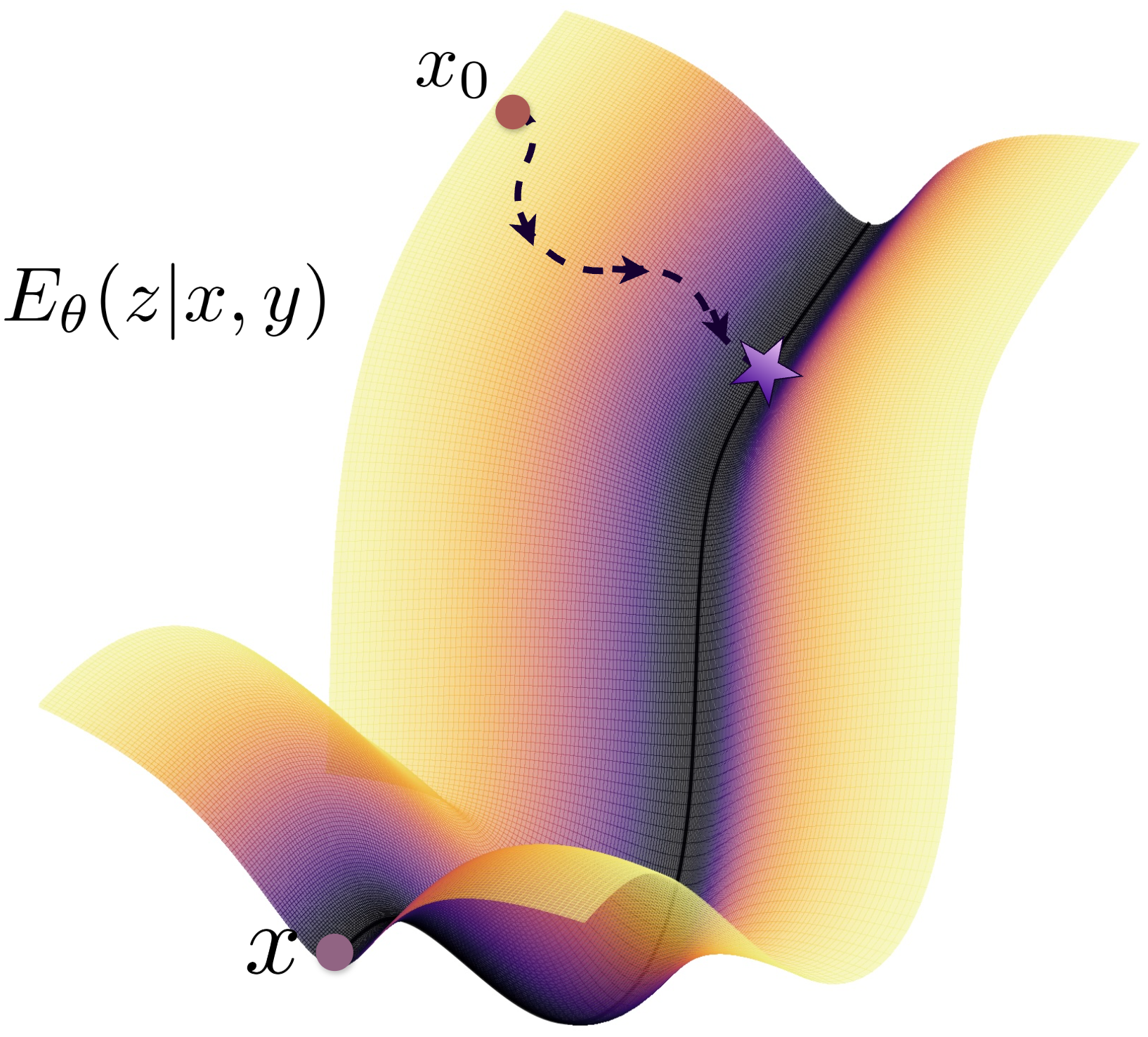}
    \caption{\textbf{EBT-Policy Diagram.} EBT-Policy functions through searching for a low energy action trajectory in cartesian or joint space ($z$) through energy minimization. Further experiments will also be updated.}
    \label{fig:energy_landscape}
\end{figure}
    
\section{Introduction}

Robots with human-level understanding, reasoning, and planning in the physical world remains as one of the cardinal objectives of AI \cite{foundation-models}. Currently, behavior cloning has re-gained traction through Diffusion Policy~\cite{dp} to help achieve this objective: instead of fitting a brittle explicit policy that maps from a robot's environmental observations to actions, the method encodes goals and observations into a shared latent space and learns the score (gradient) of the demonstration distribution \cite{ddpm, ddim, dp}. A discrete noise schedule then provides ground-truth denoising targets, turning the generation of a manipulation sequence into a reverse-diffusion process that iteratively denoises from pure noise to a feasible action trajectory.

Diffusion and flow-based policies, however, suffer from several practical challenges. A primary concern is their reliance on an externally defined noise scheduler to model data distributions \cite{ebt, equilibrium-matching}. This reliance introduces a disconnect between the model’s generative dynamics and the true distribution of robot demonstrations, often resulting in slow convergence, high computational cost during training, and inefficient inference due to the large number of denoising steps required \cite{pi0-fast, gpi, flow-policy}. Another critical issue is exposure bias \cite{exposure-bias, alleviating-exposure-bias}, an inherent flaw in their chain-structured generation process that differs from the training procedure. Errors introduced in early stages of the diffusion sequence propagate downstream and accumulate through successive steps \cite{error-propagation}, amplifying prediction inconsistencies. This cascading error effect makes diffusion-based policies particularly difficult to train or interpret in real-world robotic settings where distribution shifts are common.

Energy-Based Models (EBMs)~\cite{ebm-tutorial, implicit-generation} naturally offer promise in resolving many of the core issues challenging diffusion models, through the avoidance of relying on specific schedulers for noise, and the learning of an energy landscape that can reduce error accumulation issues during inference~\cite{ebt}. Particularly, by sampling with Markov Chain Monte Carlo (MCMC) guided by the energy, the policy repeatedly re-evaluates and corrects intermediate proposals, instead of committing to a fixed denoising trajectory, which can reduce cascading mistakes and schedule sensitivity~\cite{du2024learning, ebt}.

While theoretically attractive, Implicit Policies parametrized through EBMs have failed to become a standard for policy learning due to challenges with scalability and training stability~\cite{implicit-generation, du2020improved}. Therefore, motivated by recent work demonstrating a scalable and stable approach for training EBMs deemed Energy-Based Transformers (EBTs)~\cite{ebt}, we introduce EBT-Policy, a modern recipe for Implicit Policies based on EBTs. EBT-Policy broadly involves training EBMs with regularized losses, as opposed to the contrastive losses commonly used in existing Energy-Based Implicit Policies, the usage of modern Transformer architectures, and the usage of several energy landscape regularization techniques to better shape the energy landscape during training~\cite{ebt}. To further enhance training stability and inference efficiency, we introduce additional components including energy-scaled MCMC step sizes, pre-sample normalization, Nesterov-accelerated gradients~\cite{nesterov1983}, and scaled Langevin Dynamics. These mechanisms collectively improve training and inference stability, enabling more effective action sampling. As a result of Energy-Based training, EBT-Policy naturally enables dynamic inference computation through iterative energy minimization and uncertainty-driven behavior through the energy scalar, which we demonstrate through experiments.

Across both simulated and real-world tasks, we find training and inference with EBT-Policy to occur much faster than that of Diffusion Policy. For example, during inference, we find EBTs often can complete tasks to a high success rate with only 2 inference steps, as compared to Diffusion Policy requiring 100 steps to complete the same task at the same success rate. Further, we achieve state-of-the-art results on Squre and Tool Hang task in robomimic, outperforming Diffusion Policy by as much as 24\%. Qualitative results also suggest that EBT-Policy successfully captures uncertainty in difficult portions of robotics tasks, offering promise towards a world where robots are more aware of real-world challenges. Lastly, EBT-Policy demonstrates \textit{emergent capabilities}, where EBT-Policy learns retry behavior without any explicit training data or supervision to do so, achieving a capability that has long been sought after in Behavior Cloning~\cite{ross2010efficient}. Our results demonstrate the strong potential of Implicit policies parametrized by EBMs that act more dynamically, with uncertainty awareness, and with emergent capabilities. 
%TODO redo based on newest results, highlight best results
\section{Related Work}

\begin{figure*}[t]
  \centering
  % 0.75 or 0.80 → pick the scale that looks right for your graphic
  \includegraphics[width=0.85\textwidth]{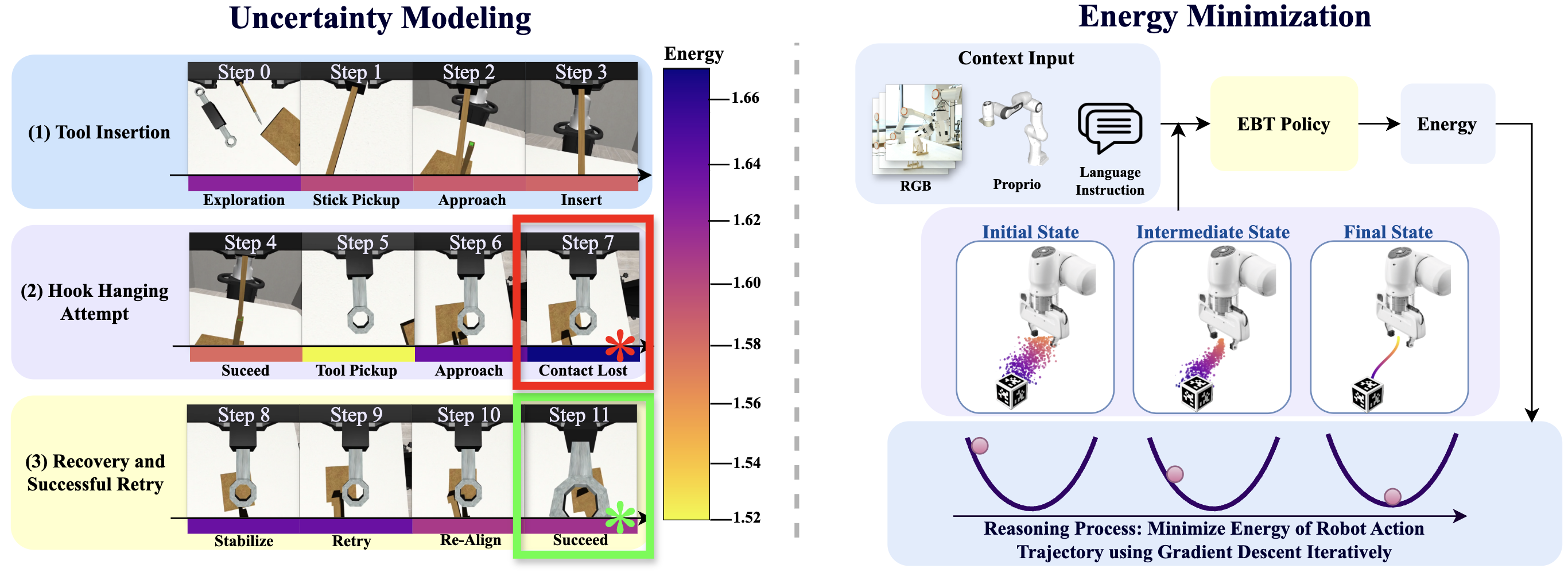}
  \caption{\textbf{Explaining Uncertainty Modeling.} 12 frames are grouped into three phases: (1) Tool Insertion, (2) Hook Hanging Attempt, and (3) Recovery \& Successful Retry. Color bar beneath each frame encodes per-frame energy predicted by the model, where a lower energy indicates a higher certainty in EBT-Policy. Notably, red (Step 7) marks the failure that triggers an EBT-Policy retry, while green (Step 11) marks the successful correction. Together, these steps highlight EBT-Policy’s interpretability and physical reasoning: using energy-based uncertainty to decide whether to continue or retry and how to adjust actions.
  \textbf{Explaining Energy Minimization.} EBT-Policy receives inputs (RGB frames, robotic proprioception, and language instructions) and assigns an energy to candidate action trajectories. Starting from a noisy initialization, the trajectory is iteratively updated by gradient descent on this energy, yielding starting states to a final executable plan. Optimization terminates when the energy converges to a minimum, as illustrated by the energy-landscape sketch.}
  \label{fig:main}
\end{figure*}

\subsection{Implicit Behavior Cloning with Diffusion}
Implicit Behavior Cloning (IBC) frames policy learning as conditional energy modeling over actions; at inference, actions are obtained by minimizing the learned energy, enabling multimodality without teacher forcing \cite{ibc}. Within Behavior Cloning (BC), however, Diffusion-based generative policies \cite{ddpm, ddim, flow-matching} have emerged as the dominant approach \cite{dp, uva, rdt, octo, robograsp, svp, crossformer, dit-policy, tenma, edp, pi0, flow-policy, gr00t-n1, fp3, unified-world-model}, which amortize the score (gradient of the energy function) prediction, rather than differentiating an underlying explicit scalar energy function. The success of Diffusion has in part been due to high training stability and capacity to model multimodal high-dimensional action distributions \cite{dp, dit-policy, rdt}. 

\subsection{Implicit Behavior Cloning with Energy-Based Models}
In contrast to diffusion-based BC, implicit Energy-Based Models (EBMs) learn an explicit energy landscape whose equilibrium dynamics define the policy \cite{ebm-tutorial, implicit-generation}. Previously, energy-based policies \cite{ibc, energy-based-il} were found to be unstable for training due to requiring an exponential number of negative samples with increasing data dimensionality~\cite{lecun2022path}. Recently, however, explicit Energy-Based methods have seen some promise in improved scalability~\cite{ebt}, which offers promise in resolving some of the challenges with Diffusion.

\subsection{Reasoning and Emergent Behavior}
Reasoning has become a hot topic within AI, with several approaches emerging, including Chain-of-Thought~\cite{chain-of-thought1}, recursive reasoning~\cite{hao2024training}, and Reinforcement Learning for reasoning~\cite{guo2025deepseek}. Within Robotics, reasoning has often been performed with Vision-Language-Action (VLA) models~\cite{rt2, openvla, opendrive-vla}, which can leverage a rich set of pretrained knowledge in Vision-Language Models to reason. The most similar works to ours involving reasoning leverage Energy-Based Models for reasoning, which have demonstrated strong generalization performance across discrete and continuous modalities~\cite{iterative-energy-minimization, energy-diffusion, ebt}.

In the robotics community, several existing works achieve emergent retry behavior~\cite{belkhale2023hydra, cideron2023get, liu2024visual}. However, these works focus on achieving retry behavior via planning, goal-conditioning, or hierarchical execution---not from a vanilla behavior-cloned policy that solely outputs instantaneous actions. To the best of our knowledge, EBT-Policy is the first approach to demonstrate emergent retry behavior \textit{solely from action prediction} and without training data for retries when performing behavior cloning.

\section{EBT-Policy Formulation}

We formulate visuomotor control policies as \textbf{Energy-Based Transformers (EBTs)}, which we refer to as \textbf{EBT-Policy}. EBT-Policies represent complex, multi-modal, action distributions through an energy landscape spanning the action space, framing energy-based learning and inference as a continuous optimization/sampling process. This formulation enables flexible and implicit modeling of unnormalized action likelihoods, while maintaining stability and scalability across high-dimensional visuomotor tasks.

\subsection{Energy-Based Transformers}

Energy-Based Transformers (EBTs) \cite{ebt} belong to the broader class of \emph{Energy-Based Models (EBMs)} \cite{ebm-tutorial}, which learn to map inputs to a scalar energy, often referred to as the energy function. Particularly, EBMs learn a probability distribution parametrized in the form of a Boltzmann distribution $p_\theta(x) = \frac{e^{-E_{\theta}(x)}}{Z(\theta)}$. In this work, we focus on unnormalized EBMs, which forgo the intractable partition function $Z(\theta)$, in exchange for modeling unnormalized probabilities~\cite{ebt}. In this approach, the energy function quantifies the model’s internally inferred compatibility between an input and candidate predictions, and the generation problem corresponds to minimizing this energy, often by gradient descent. Intuitively, the energy scalar can be interpreted as a compatibility estimate between the input variables in the forward pass, where the energy minimization process on the backward pass can be interpreted as a series of optimizations by minimizing this scalar energy.
% to lead to a state of certainty of its current inputs and its intended decision. Minimizing energy thus corresponds to maximizing the model’s internal certainty about the correctness of a predicted outcome given an observation.

EBTs usually generate predictions via iterative energy minimization using gradient descent, which can be expressed as:
\[
  \hat{\mathbf{y}}_{i+1} = \hat{\mathbf{y}}_{i} - \alpha_i \nabla_{\hat{\mathbf{y}}} E_\theta(\mathbf{x}, \hat{\mathbf{y}}_{i}) + \eta_i
\]
where $E_\theta(\mathbf{x}, \hat{\mathbf{y}}_{i})$ denotes the predicted energy for a candidate prediction $\hat{\mathbf{y}}_{i}$ given input $\mathbf{x}$, $\alpha_i$ as the energy-scaled step size, and Langevin Dynamics noise $\eta_i$. This sampling procedure is approximately equivalent to a Markov Chain Monte Carlo (MCMC) method known as Langevin Dynamics, in which the prediction $\hat{\mathbf{y}}_{i+1}$ is updated via gradient descent with respect to the scalar energy function and some noise. We iterate the number of inference steps adaptively according to the current energy level, and scale $\alpha$ as a function of the energy to enable stable convergence and high flexibility.

\subsection{EBTs for Robot Learning}

The objective of EBT-Policy is to generate a sequence of robot actions---either in joint space or Cartesian space---conditioned on recent multimodal sensory observations and, optionally, natural language instructions. Formally, let $\mathcal{L}$ denote the space of language commands, $\mathcal{X}$ the space of RGB visual observations, and $\mathcal{Z}$ the space of proprioceptive states. At each time step $t \in \mathbb{N}^+$, the input to the policy is defined as:
\[
(\ell, o_t) \in \mathcal{L} \times \left(\mathcal{X}^h \times \mathcal{Z}^h\right),
\]
where $\ell$ represents a natural language instruction and $o_t = (X_{t-h+1:t}, z_{t-h+1:t})$ denotes an $h$-step observation window, with $X_{t-h+1:t} \in \mathcal{X}^h$ corresponding to the sequence of RGB frames and $z_{t-h+1:t} \in \mathcal{Z}^h$ corresponding to proprioceptive state histories.

The policy $\pi$ maps these inputs to a sequence of future actions over a prediction horizon $n$
\[
\pi: \mathcal{L} \times \mathcal{X}^h \times \mathcal{Z}^h \rightarrow \mathcal{A}^n
\]
where $\mathcal{A}^n$ denotes the continuous action space. The output sequence $a_{t:t+n-1} = (a_t, a_{t+1}, \ldots, a_{t+n-1})$ corresponds to the robot's planned actions for the next $n$ steps, aimed at achieving the task described by $\ell$.

In the EBT formulation, the policy is modeled as an \emph{energy minimization process} rather than learning an explicit policy \cite{ibc, ebm-tutorial}, which is defined as:

$$\pi: \text{argmin}_{a \in \mathcal{A}^n} E_\theta(\ell, o_t, a)$$

where the energy function is learned such that it assigns low energy to action trajectories consistent with the given sensory observations and language instruction. 
% The global minimum of this energy landscape represents the optimal action sequence—the most compatible under the current context. %unfortunately this isnt necessarily true so we need to be careful saying it
During inference, the EBT iteratively refines sampled actions via gradient-based updates to descend toward a low-energy region, effectively sampling from the manifold of task-consistent action sequences.

\subsection{Diffusion vs. Energy-Based Policies}

While both \textbf{Diffusion Policy} and \textbf{EBT-Policy} generate actions through iterative refinement, they differ fundamentally in their probabilistic formulation and inference dynamics. DPs are score-based models that represent the data distribution via a stochastic denoising process, where a network $\epsilon_\theta(\mathbf{a}_t, t, \ell, o_t)$ learns to reverse a forward noise process. Through denoising score matching, this network implicitly estimates the gradient of the conditional log-likelihood,
\[
\nabla_{\mathbf{a}} \log p(\mathbf{a} \mid \ell, o_t)
\]
which corresponds to the direction of increasing probability under the true data distribution. This formulation avoids the intractable partition function in classical energy-based models by directly modeling the score field, resulting in stable training and well-behaved likelihood gradients \cite{dp}.

In contrast, EBTs define an unnormalized energy function $E_\theta(\ell, o_t, \mathbf{a})$ whose negative gradient yields the same score:
\[
\nabla_{\mathbf{a}} \log p(\mathbf{a} \mid \ell, o_t) = -\nabla_{\mathbf{a}} E_\theta(\ell, o_t, \mathbf{a})
\]
Rather than learning to denoise stochastic samples, EBTs learn this energy landscape directly and infer actions by descending it toward low-energy (high-likelihood) states. Conceptually, diffusion models approximate the gradient of the energy, while EBTs explicitly learn the energy itself---making EBTs explicitly model the underlying data density with equilibrium dynamics.

\subsection{Why EBTs Work for Robot Policies}
\smallskip
\noindent\textbf{1. Equilibrium dynamics.}
EBTs learn a single, time-invariant explicit energy landscape which enables dynamics where OOD data points are brought closer to the data manifold. These dynamics effectively reduce exposure bias and compounding error, which commonly haunt Diffusion/Flow models once they start to drift OOD~\cite{ebt}. Additionally, we find that these dynamics lead to far fewer inference steps, faster training than diffusion/flow models as confirmed in Figure~\ref{fig:success_vs_epochs}. We believe this is due to the non-reliance on a noise schedule, time-variance, and ODE solvers, which impose strict constraints on model representations and the path from noise to data. Rather, EBTs can take whatever path from noise to data works best, as paths are learned end-to-end rather than being based on a fixed schedule. 
% as the policy is directly learning the data distributions of the robot demonstration data.

\smallskip
% TODO maybe eventually add this back?
% \noindent\textbf{2. High-dimensional observations = built-in stability.}
% Continuous, high-cardinality sensor streams (RGB + proprio) provides flexibility to the plausible action manifold, restricting the complexity of the energy landscape and EBTs gradient-based energy minimization process eliminates the exponential negative-mining problem faced by contrastive IBC \cite{dp, ibc, ebt}.

\smallskip
\noindent\textbf{2. Uncertainty modeling.}
Energy landscapes can succesfully capture uncertainty \cite{ebt, Dawid_2024}, where uncertainty can be represented as high energy or landscapes with several local minima. Consqeuently, EBTs naturally enable uncertainty-aware sampling, where sampling continues until the energy has converged (or the gradient goes below a certain norm threshold); this results in hard states getting more gradient steps, and easy states getting less, resulting in interpretable, compute-adaptive behaviour, which we verify in Figure~\ref{fig:main}.

\smallskip
\noindent\textbf{3. OOD robustness via discriminative learning.}
The same scalar energy acts as a verifier: low energy $\Rightarrow$ plausible, high energy $\Rightarrow$ reject. This built-in discriminator makes EBTs noticeably less brittle to environmental changes than generative diffusion/flow policies, as we verify in Figure~\ref{fig:dp_edp_compare}.

\begin{figure}[t]
    \centering
    % ---- First row ----
    \begin{subfigure}[t]{0.45\columnwidth}
        \centering
        \includegraphics[width=\linewidth]{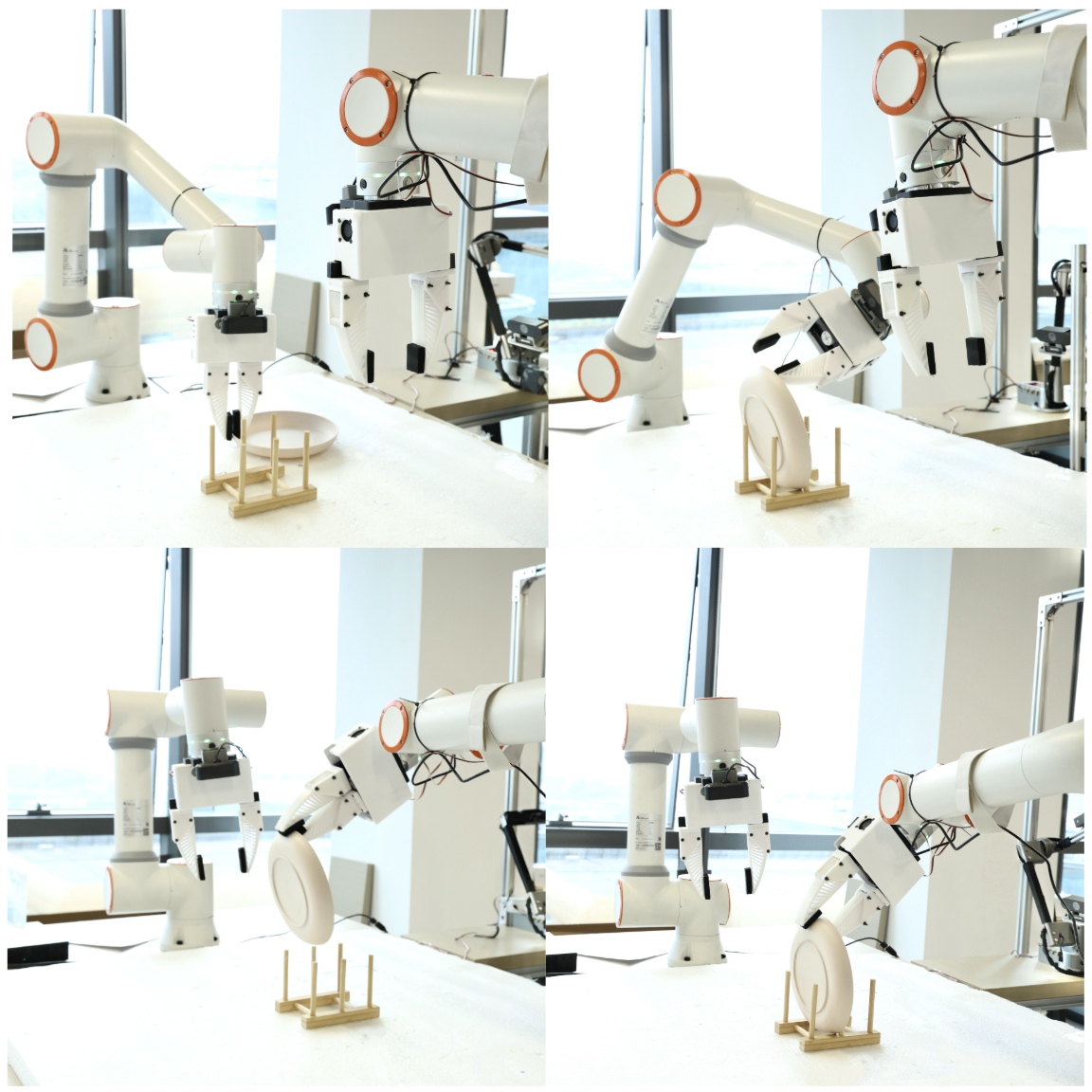}
        \caption{Collect Dish}
    \end{subfigure}
    \hfill
    \begin{subfigure}[t]{0.45\columnwidth}
        \centering
        \includegraphics[width=\linewidth]{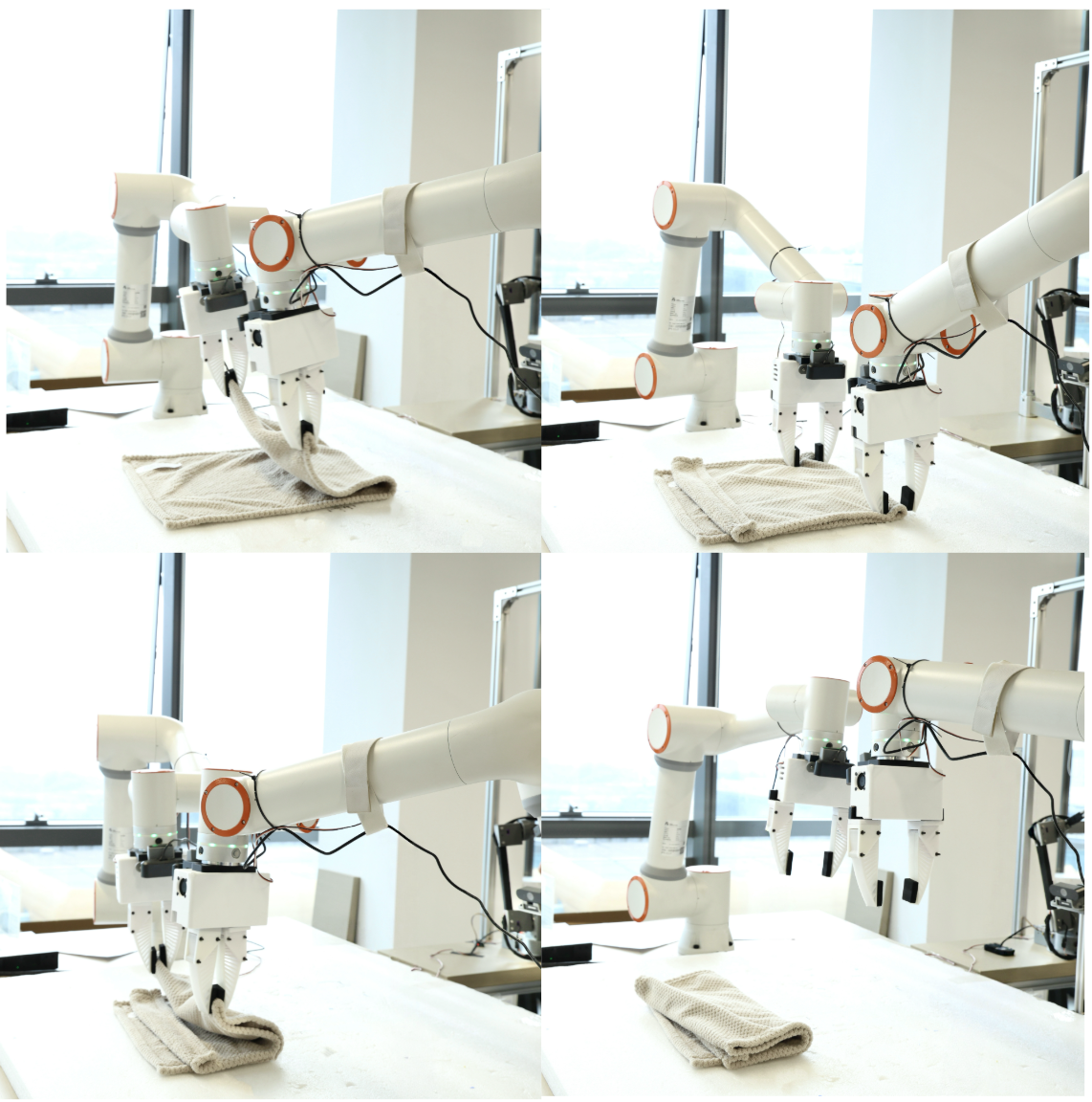}
        \caption{Fold Towel}
    \end{subfigure}
    
    \vspace{4pt} % space between rows
    
    % ---- Second row ----
    \begin{subfigure}[t]{\columnwidth}
        \centering
        \includegraphics[width=\linewidth]{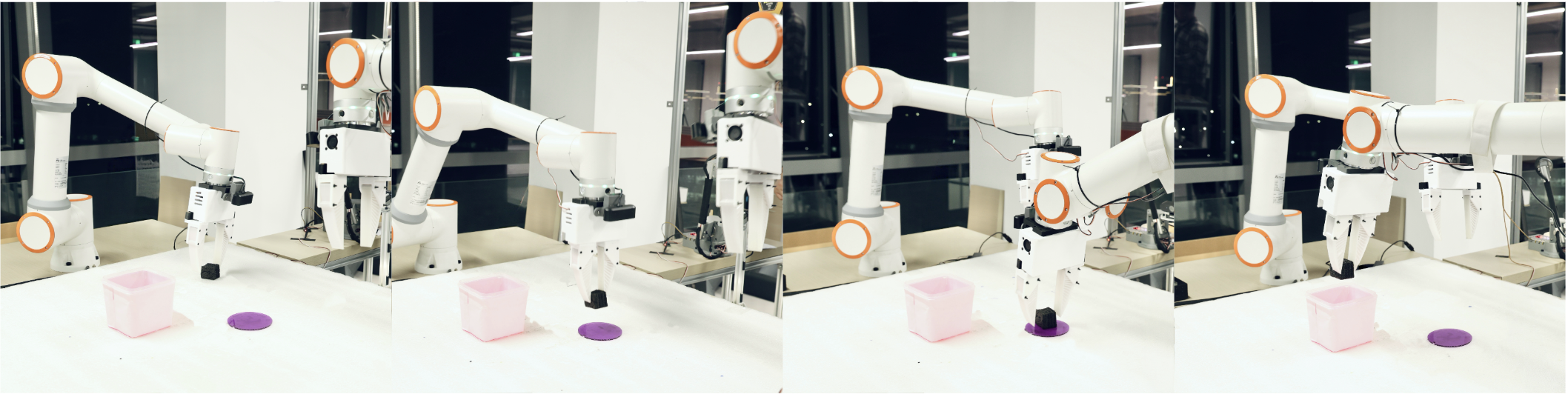}
        \caption{Pick and Place}
    \end{subfigure}
    
    \caption{\textbf{Demonstrations from tabletop, real-world tasks.}}
    \label{fig:real_tasks}
\end{figure}

\subsection{Towards Embodied Reasoning in Robotics}

We claim that EBTs offer a unified computational framework for \emph{embodied System 2 reasoning} in robots \cite{kahneman2011thinking, ebt}. Unlike Vision-Language-Action (VLA) \cite{rt2, openvla, pi0, pi_05, gr00t-n1} architectures and diffusion-based policies \cite{rdt, dit-policy, dp, edp, fp3, flow-policy}, which rely on discrete tokenization, fixed compute budgets, and/or externally imposed reasoning modules, EBTs integrate perception, reasoning, and control within a single energy-minimization objective. This formulation enables robots to reason natively in continuous sensory space and to allocate computation adaptively based on internal uncertainty. 

Three properties characterize this native reasoning capability:

\smallskip
\noindent\textbf{1. Continuous-native sensor grounding.} EBTs operate directly on continuous sensory representations (e.g., proprioceptive and camera data), enabling reasoning in the robot’s native embodiment space rather than over discrete space (e.g., written text).

\smallskip
\noindent\textbf{2. Uncertainty-aware dynamic inference.} The number of inference steps is dynamically allocated according to task complexity and internal energy gradients, allowing EBTs to modulate computational effort based on internal uncertainty in observation and goal data.

\smallskip
\noindent\textbf{3. Self-supervised cognitive optimization.} Training is driven by internal energy dissipation-minimizing energy for consistent action trajectories, providing a practical solution to a self-supervised System 2 training procedure.

\subsection{EBT-Policy Ingredients}
Training energy-based trajectory (EBT) policies is hindered by two issues: action demonstration data is inherently multimodal \cite{dp, rdt}, tempting the model into local energy minima, and the iterative MCMC refinement creates long gradient chains that can often explode in practice, much like the well-documented issues of traditional recurrent-style models \cite{exploding-gradients}. We introduce design tweaks that curb both problems and stabilize learning.

\begin{table}[t]
\centering
\renewcommand{\arraystretch}{1.2}
\setlength{\tabcolsep}{6pt}
\begin{tabular}{ll}
\toprule
\textbf{EBT-Policy Hyperparameters} & \textbf{Value} \\
\midrule
Base step size ($\eta_b$) & 1000 \\
Step size scaling factor ($c$) & 1.5 \\
Min. LD scale ($\sigma_{\min}$) & 0.002 \\
Max. LD scale ($\sigma_{\max}$) & 0.2 \\
Num. of base MCMC train steps & 6 \\
Num. of randomized additional steps & 3 \\
Max. MCMC inference steps & 20 \\
Gradient clipping norm & 1.0 \\
\bottomrule
\end{tabular}
\caption{\textbf{Training hyperparameters for EBT-Policy.}}
\label{tab:ebt_hyperparams}
\end{table}

\subsubsection{Learning Multimodal Action Distributions}
We introduce several \textbf{training-only} mechanisms that enable the EBT-Policy to capture and explore multi-modal action distributions.

\noindent
\textbf{Randomized Sampling Steps.}
The total number of \emph{Markov Chain Monte Carlo} (MCMC) sampling steps is randomized to promote stochastic exploration across different energy modes.

\smallskip
\noindent
\textbf{Scaled Langevin Dynamics.}
Each sampling step is augmented with \emph{scaled Langevin Dynamics} \cite{bayes-learning}, which injects controlled stochasticity into the sampling process. The magnitude of the injected noise follows a \emph{cosine Langevin Dynamics annealing} schedule \cite{cosine-annealing}, applied to the noise standard deviation such that $\mathcal{N}(0, \sigma_t)$. The noise level smoothly decays from $\sigma_{\max}$ to $\sigma_{\min}$ over $T$ total steps:
\begin{equation}
\sigma_t = \sigma_{\min} + \tfrac{1}{2} \left( \sigma_{\max} - \sigma_{\min} \right)
\left[ 1 + \cos\!\left( \frac{\pi \cdot t}{T} \right) \right]
\label{eq:cosine_sigma}
\end{equation}
High-energy regions therefore receive stronger noise for broad exploration, while low-energy regions use smaller noise for precise convergence.
% TODO discuss exploration with higher noise -> lower noise exploitation?

\smallskip
\noindent
\textbf{Step Size Randomization.}
We use \emph{MCMC step size randomization} by sampling the initial step size as $\eta \sim \mathcal{N}(\eta_b / c,\, \eta_b \cdot c)$, where $\eta_b$ denotes the base step size and $c$ is a scaling factor controlling the mean and variance. This stochasticity encourages diverse sampling trajectories and improves robustness against local minima. Moreover, the base step size $\eta_b$ contributes to training stability by inversely scaling with the predicted gradient magnitudes, thereby mitigating the risk of exploding gradients.

\smallskip
\noindent
\textbf{Nesterov Acceleration.}
Finally, we employ \emph{Nesterov’s accelerated gradients} \cite{nesterov1983} to enable smoother traversal of complex energy surfaces and to help escape shallow local minima.

\subsubsection{Model Stability}
To address training instability and prevent excessive parameter updates, we incorporate several stabilization mechanisms that collectively ensure smooth and reliable optimization of EBT Policies.

\noindent
\textbf{Energy-Scaled Step Sizes.}
To prevent overshooting or undershooting minima, we employ \emph{energy-scaled step sizes}, defined as:
\begin{equation}
\alpha_i = \eta \, \exp\big(E_{\theta}(\mathbf{x}, \hat{\mathbf{y}}_i)\big)
\label{eq:energy_scaling}
\end{equation}
which adaptively modulate update magnitudes in proportion to predicted energy, yielding well-conditioned optimization.

\smallskip
\noindent
\textbf{Pre-Sample Normalization.}
Action trajectories are normalized to $[-1, 1]$ prior to input into the EBT, preventing uncontrolled growth of action magnitudes and gradients during sampling. For this paper, we use RMSNorm for pre-sample normalization \cite{rmsnorm}.

\smallskip
\noindent
\textbf{Gradient Clipping.}
To mitigate exploding gradients causing instability and prevent large updates from saturating model weights, we apply \emph{gradient clipping} \cite{grad-clip}, constraining the global gradient norm to a maximum value of 1.0. Empirically, we find this technique to be the most critical component for ensuring stable and reliable training of EBT Policies.

\subsection{EBT-Policy Training}
Algorithm~\ref{alg:train} outlines the training procedure for the EBT Policy. During each training iteration, the target trajectory is initialized as pure noise. The model then encodes its conditional inputs, including multimodal sensory observations and, when available, language instructions. Subsequently, randomized Markov Chain Monte Carlo (MCMC) step sampling is performed, where trajectories are normalized, and Langevin Dynamics are injected according to an annealed noise schedule $\sigma_i$. The model predicts the energy landscape $E_\theta$, after which the step size is randomized and scaled proportionally to the predicted energy. The loss is then accumulated over each sampled trajectory by comparing the denoised trajectory with the ground-truth demonstration.

\begin{algorithm}
\caption{EBT-Policy Training}
\begin{algorithmic}[1]
\Require Context $x$, Target $y$, Context encoder $f_\theta(x)$, EBT decoder $E_\theta(z,\hat y)$
\Ensure MCMC Steps $N$
  \State Sample $\hat{\mathbf y}_0 \sim \mathcal N(0,I)$
  \State $z \gets f_\theta(x)$
  \State $\mathcal L \gets 0$
  \For{$i = 0$ \textbf{to} $N-1$}
    \State $\hat{y}_i \gets \text{RMSNorm}(\hat{y}_i) + \mathcal{N}(0, \sigma_i)$ 
    \State $\hat{y}_{i+1} \gets \hat{y}_{i} - \alpha_i \nabla_{\hat{y}} E_\theta(z, \hat{ y}_{i})$
    \State $\mathcal L \gets \mathcal L + \text{MSE}(\hat{y}_{i+1}, y)$
  \EndFor
  \State update $E_\theta$, return $\mathcal L$
\end{algorithmic}
\label{alg:train}
\end{algorithm}

\subsection{Dynamic Inference in EBT-Policy}

Algorithm~\ref{alg:inference} describes the dynamic inference process of the EBT-Policy. Unlike conventional diffusion- or sampling-based policies that use a fixed number of inference steps, EBT-Policy performs \textbf{adaptive energy descent}, dynamically determining the number of Markov Chain Monte Carlo (MCMC) updates required for convergence. The process continues until either the maximum step limit $N$ is reached or the energy gradient norm $\| g \|_2$ falls below a threshold $\tau$.

This \textbf{dynamic termination criterion} allows EBT-Policy to automatically allocate computation based on its internal \emph{certainty state}—requiring fewer steps for low-energy (high-certainty) regions and more steps for uncertain states. This makes inference both computationally efficient and confidence-aware, enabling EBTs to adaptively balance precision and runtime without manual tuning.

\begin{algorithm}
\caption{EBT-Policy Dynamic Inference}
\begin{algorithmic}[1]
\Require Context $x$, Context encoder $f_\theta(x)$, EBT decoder $E_\theta(x,\hat y)$
\Ensure Max MCMC steps $N$, Gradient cutoff $\tau$
  \State Sample $\hat{\mathbf y}_0 \sim \mathcal N(0,I)$
  \State $z \gets f_\theta(x)$
  \State $i \gets 0$ 
  \State $g \gets \infty$
  \While{$i < N$ \textbf{and} $\| g \|_2 > \tau$}
    \State $\hat{y}_i \gets \text{RMSNorm}(\hat{y}_i)$
    \State $g \gets \nabla_{\hat{y}} E_\theta (z, \hat{y}_i)$
    \State $\hat{y}_{i+1} \gets \hat{y}_{i} - \alpha_i \cdot g$
    \State $i \gets i + 1$
  \EndWhile
\State return $\hat{y}_M$
\end{algorithmic}
\label{alg:inference}
\end{algorithm}
\section{Experiments}

To evaluate the proposed \emph{EBT-Policy}, we conduct a range of experiments across both simulated benchmarks and physical robotic environments. Two architectural variants of EBT-Policy are considered, as shown in Table~\ref{tab:ebt_variants}.

\begin{table}[t]
\centering
\renewcommand{\arraystretch}{1.25}
\setlength{\tabcolsep}{6pt}
\begin{tabular}{lcc}
\toprule
\textbf{Property} & \textbf{EBT-Policy-S} & \textbf{EBT-Policy-R} \\
\midrule
Size & $\sim$30M & $\sim$100M \\
Task & Simulation & Real World \\
Language Encoder & N/A & T5-S~\cite{t5} \\
Vision Encoder & ResNet-18~\cite{resnet} & DINOv3-S~\cite{dinov3} \\
\bottomrule
\end{tabular}
\caption{\textbf{Comparison of EBT-Policy variants.} 
EBT-Policy-S is a compact Transformer used for controlled simulation studies, 
while EBT-Policy-R is a larger multimodal variant designed for real-world, language-conditioned, and multitask policy learning.}
\label{tab:ebt_variants}
\end{table}

The distinction between the two variants of EBT-Policy serve distinct experimental goals:

\smallskip
\noindent\textbf{1. Controlled simulation benchmarks.} 
\emph{EBT-Policy-S} adopts the same architecture as \emph{Diffusion Policy} to enable a fair, one-to-one comparison under identical simulation conditions, isolating the effect of energy-based versus diffusion-based objectives. 

\smallskip
\noindent\textbf{2. Real-world robotic manipulation evaluation.} 
\emph{EBT-Policy-R} is deployed in physical environments and compared with state-of-the-art diffusion-based policies to assess scalability and real-world applicability, under a fixed $\sim$100M parameter budget for fairness. Comparisons with billion-parameter Vision-Language-Action or robot foundation models are omitted, as their internet-scale pretraining obscures a fair comparison of policy objectives. All baselines are diffusion transformers of comparable size ($\leq$150M) trained on the same robot-demonstration data, providing a controlled evaluation of the energy-based formulation.

\subsection{Task Design}

\textbf{Simulation Tasks.} 
We first evaluate in simulation using the robomimic benchmark suite~\cite{robomimic}, focusing on four manipulation tasks of increasing difficulty: \emph{Lift}, \emph{Can}, \emph{Square}, and \emph{Tool Hang}. 
Each task in robomimic is characterized along several axes: task difficulty, positional variance, goal precision, object dynamics, and temporal horizon-designed to progressively challenge policy robustness and generalization.
The robomimic environment provides a standardized and reproducible setting for evaluating robot policy performance.

% TODO keep this commented?
% \begin{table}[t]
% \centering
% \renewcommand{\arraystretch}{1.25}
% \setlength{\tabcolsep}{5pt}
% \begin{tabular}{lcccc}
% \toprule
% \textbf{Property} & \textbf{Lift} & \textbf{Can} & \textbf{Square} & \textbf{Tool Hang} \\
% \midrule
% Task Difficulty & 1 & 2 & 3 & 5 \\
% Positional Variance & 1 & 3 & 4 & 1 \\
% Goal Precision & 1 & 2 & 4 & 5 \\
% Object Dynamics & 1 & 1 & 1 & 3 \\
% Task Horizon Length & 1 & 2 & 2 & 4 \\
% \bottomrule
% \end{tabular}
% \caption{\textbf{Simulated-task complexity profile.}
% Scores (1–5) reflect increasing challenge across task difficulty, object position variance, goal precision, object dynamics, and task horizon length.
% From \emph{Lift} to \emph{Tool Hang}, task complexity rises progressively, providing a quantitative curriculum for evaluating policy scalability.}
% \label{tab:task_properties}
% \end{table}

\begin{figure}[t]
    \centering
    \begin{subfigure}[t]{0.242\columnwidth}
        \centering
        \includegraphics[width=\linewidth]{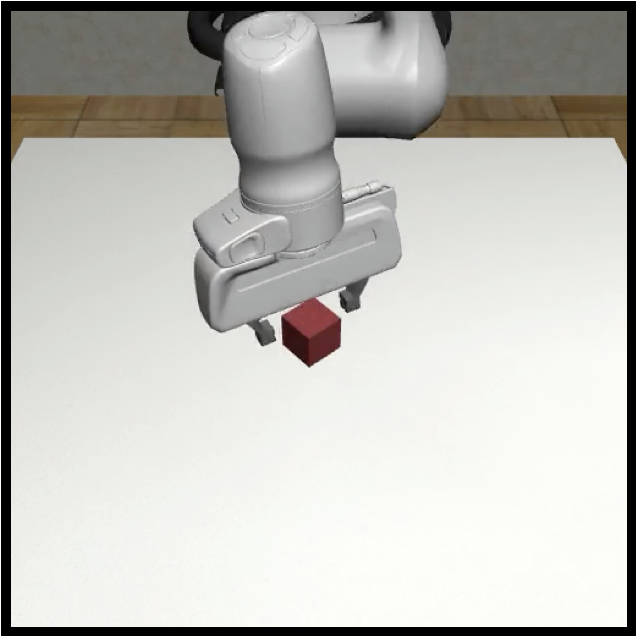}
        \caption{Lift}
    \end{subfigure}
    \begin{subfigure}[t]{0.242\columnwidth}
        \centering
        \includegraphics[width=\linewidth]{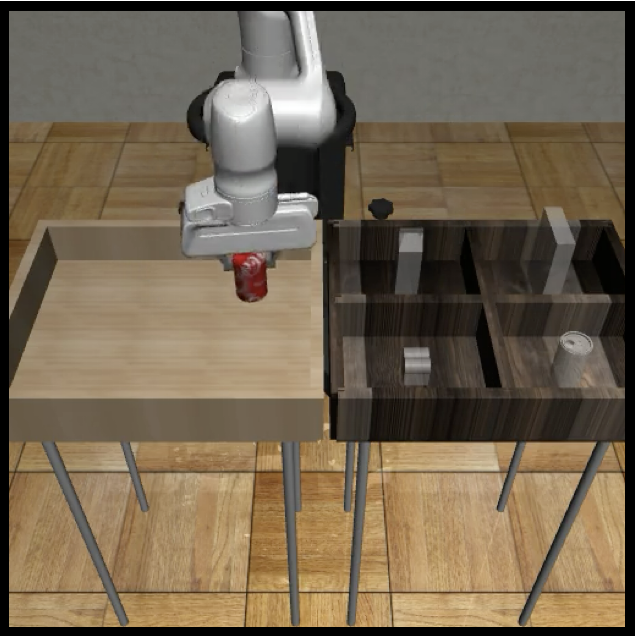}
        \caption{Can}
    \end{subfigure}
    \begin{subfigure}[t]{0.242\columnwidth}
        \centering
        \includegraphics[width=\linewidth]{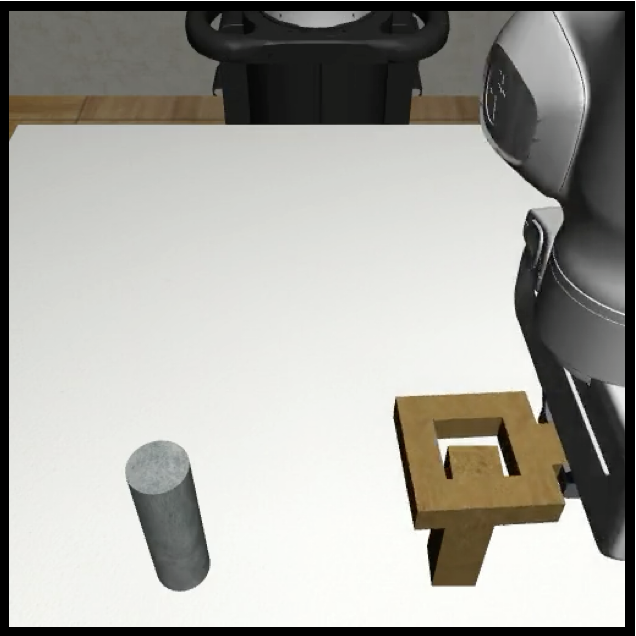}
        \caption{Square}
    \end{subfigure}
    \begin{subfigure}[t]{0.242\columnwidth}
        \centering
        \includegraphics[width=\linewidth]{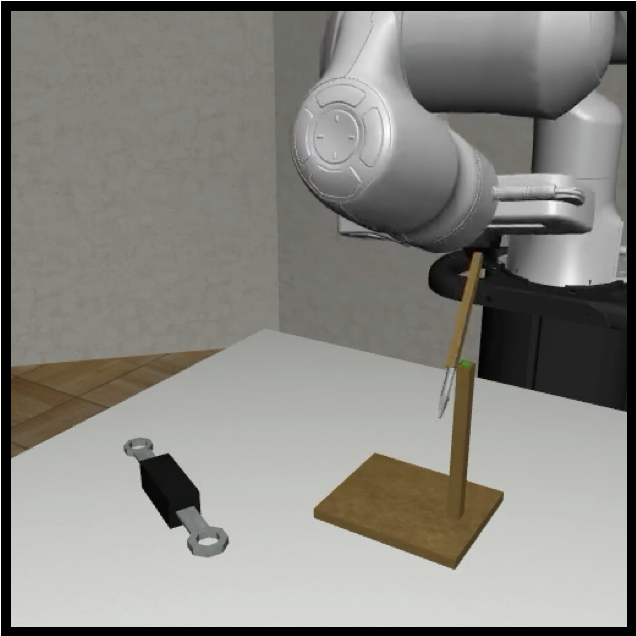}
        \caption{Tool Hang}
    \end{subfigure}
    \caption{\textbf{Representative tasks in robomimic~\cite{robomimic}.}}
    \label{fig:robomimic_tasks}
\end{figure}

\smallskip
\noindent\textbf{Real-World Tasks.} 
To assess real-world transfer, we design a set of general manipulation tasks on a dual-arm robotic platform.
These tasks require multi-task learning, robustness to real-world sensor noise, and goal understanding:

\begin{itemize}
    \item \textbf{FoldTowel:} Both arms collaboratively fold a towel twice, testing deformable-object manipulation and robustness to initial-state variability.
    \item \textbf{PlacePan:} The arm grasps a pan and places it on a rack at varying positions and orientations, evaluating spatial generalization and placement precision.
    \item \textbf{PickAndPlace:} One arm grasps a black block, placing it on a purple circle, for another arm to pick up the black block and put it in a pink container.
\end{itemize}

%To further evaluate robustness, we introduce out-of-distribution (OOD) variations during both training and inference:
%\begin{itemize}
%    \item \textbf{Object-Shift}: Target objects are replaced with visually and physically similar yet distinct counterparts (e.g., different towel materials or pan geometries).
%    \item \textbf{Environment-Shift}: The surrounding scene is perturbed through changes in background layout, lighting, or distractor objects.
%\end{itemize}
These tasks provide a systematic assessment of EBT’s ability to sustain stable performance under realistic environmental variability.

\subsection{Experiment Setup}
We also compare the proposed EBT-Policy with Diffusion Policy \cite{dp}. Diffusion Policy acting as the primary baseline, enabling a direct comparison between diffusion-based and energy-based policy learning. All models are trained on the same dataset, with default parameters provided in open-source implementations used.

For simulation benchmarks both \emph{Diffusion Policy} and EBT-Policy-R share identical architectures, hyperparameters, and augmentations, differing only in timestep conditioning and training objective. For Diffusion Policy, we additionally train 10- and 100-step variants to analyze the relationship between inference steps and task success.

Simulation experiments use the open-source robomimic datasets, with 50 episodes tested for success rate per simulation rollout. Real-world data is collected by us via a teleoperation system for a dual-arm, 4 RGB camera, tabletop setup.

\section{Results}

\begin{figure}[t]
    \centering
    \includegraphics[width=\columnwidth]{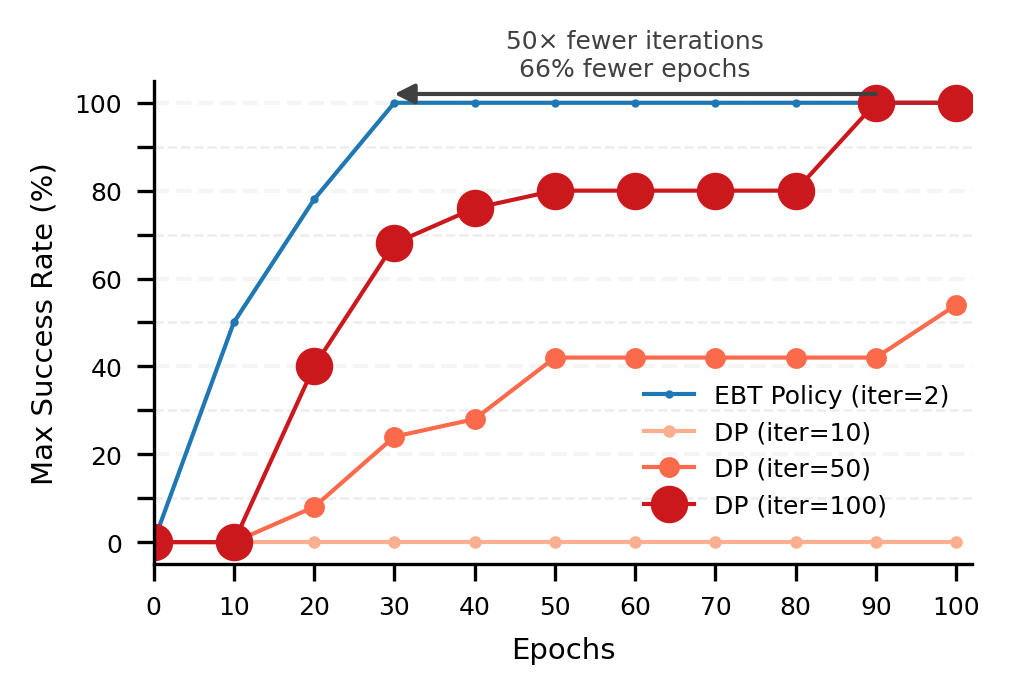}
    \caption{\textbf{Success Rates During Training.} 
    EBT-Policy exhibits rapid performance improvement, reaching $100\%$ success by epoch 30, using just 2 iterations for predicting actions. Diffusion Policy (DP), on the other hand, only reaches a $100\%$ success rate after $90$ epochs, and uses $50$ times more steps than EBT-Policy at inference, demonstrating how EBT-Policy is more efficient than DP during both training and inference.}
    \label{fig:success_vs_epochs}
\end{figure}

% In your preamble:
Simulation results are shown in Table \ref{tab:sim_results}, as shown in our results, our model outperforms Diffusion Policy on all four tasks. As shown in the results, Diffusion Policy was found to only perform well when the number of iterations was as high as 100 inference steps, which is often tenfold larger than the number of iteration steps required for EBT-Policy. This demonstrates that the compute resources for EBT-Policy during inference is much more economic than for Diffusion Policy.

As illustrated in Figure~\ref{fig:success_vs_epochs}, the EBT-policy demonstrates superior success rates and training efficiency compared to the diffusion policy (DP). Specifically, the EBT-policy achieves a 100\% success rate within just 30 epochs. In contrast, even with up to 100 iterations, the highest-performing DP variant fails to exceed a 60\% success rate over 100 epochs. This marked difference underscores the advantages of the verification-based approach in terms of both learning speed and final performance outcomes.

\begin{table}[t]
\centering
\footnotesize
\setlength{\tabcolsep}{4pt}
\renewcommand{\arraystretch}{1.15}
\begin{tabular*}{\columnwidth}{@{\extracolsep{\fill}} l c c c}
\toprule
\textbf{Real-World Task} & \textbf{Fold Towel} & \textbf{Collect Pan} & \textbf{Pick And Place} \\
\midrule
D.P. \cite{dp} & 10 & 65 & 90 \\
\textbf{EBT-Policy-R} & \textbf{86} & \textbf{75} & \textbf{92}  \\
\bottomrule
\end{tabular*}
\caption{\textbf{Real-World task success rates (\%).} Comparison of EBT-Policy-R and DP in success rates for three tasks. Numbers in bold highlights EBT-Policy-R outperforms DP in all three tasks.}
\label{tab:real_results}
\end{table}

Noteworthy is the EBT-policy's computational efficiency; it requires approximately 50 times fewer iterations and 55\% fewer epochs than the DP to achieve comparable results. This indicates not only its robustness during training but also its ability to converge more quickly by focusing on plausible action sequences rather than generating them from scratch. 
%Consequently, energy-based verification emerges as a promising alternative to generative models for robotic control tasks.

Furthermore, reducing the iteration parameter significantly degrades the DP's performance. When set to 10 iterations, the DP's success rate plummets to 0\%, starkly contrasting with the EBT-policy, which maintains high performance even at 2 iterations. This comparison vividly illustrates the EBT-policy's remarkable efficiency and reliability.

\begin{table}[t]
\centering
\footnotesize
\setlength{\tabcolsep}{4pt}
\renewcommand{\arraystretch}{1.15}
\begin{tabular*}{\columnwidth}{@{\extracolsep{\fill}} l c c c c c}
\toprule
\textbf{Benchmark Task} & \textbf{Lift} & \textbf{Can} & \textbf{Square} & \textbf{Tool Hang} \\
\midrule
D.P. \cite{dp} (n=10) & 0 & 0 & 0 & 0 \\
D.P. (n=100) & \textbf{100} & \textbf{100} & 92 & 44 \\
\textbf{EBT-Policy-S} & \textbf{100} & \textbf{100} & \textbf{98} & \textbf{68} \\
\bottomrule
\end{tabular*}
\caption{\textbf{Simulated task maximum success rates (\%).} Performance results on four robomimic tasks. SR in bold of our \textbf{EBT-Policy-S} still exceed DP's in simulation benchmarks with both fewer (10) steps and more (100) step.}
\label{tab:sim_results}
\end{table}

\FloatBarrier

\begin{figure}[!htbp]
  \centering
  \includegraphics[width=\columnwidth]{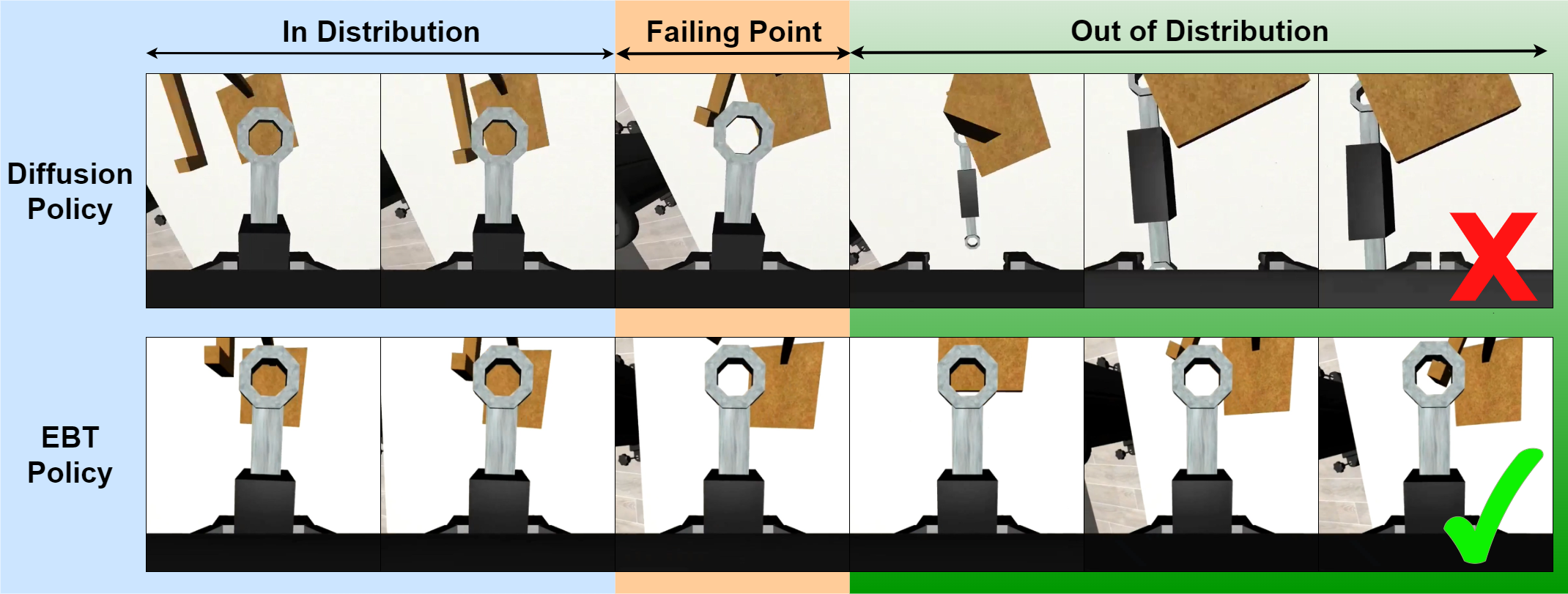}
  \caption{\textbf{EBT-Policy Emergent Retry Behavior.} A diagram illustrating the different outcomes of Diffusion Policy and EBT-Policy after encountering a failure causing covariate shift. When Diffusion Policy drifts OOD (e.g., by missing the hook), it fails catastrophically, causing compounding error and eventually divergence. EBT-Policy, after failing to hook the first time, is able to recover, and retry hooking despite being in an OOD state. This is particularly challenging as EBT-Policy has never seen such data or been explicitly trained to recover from such failed states. We hypothesize this emergent retry behavior is due to the equilibrium dynamics of the energy function, which demonstrates the emergent reasoning capabilities of EBT-Policy.}
  \label{fig:dp_edp_compare}
\end{figure}

\FloatBarrier
\section{Discussion}

A core finding in our experiments is \textbf{EBT-Policy’s emergent retry behavior} observed in the Tool Hang simulation task (Figure~\ref{fig:dp_edp_compare}). When the hook is displaced from in-distribution configurations, EBT-Policy adaptively formulates new recovery strategies, maintaining stable control and ultimately completing the task. In contrast, Diffusion Policy (DP) collapses under similar out-of-distribution (OOD) perturbations, releasing the tool and failing to recover. Both policies initially perform well during the in-distribution phase, successfully grasping and positioning the tool, but when minor contact forces cause the hook to rotate into a novel configuration, DP loses stability, whereas EBT repositions the tool and completes the placement.

To the best of our knowledge, such emergent retry behavior has not previously been observed in vanilla behavior cloning (BC) policies, which are typically brittle under OOD conditions. This distinction aligns with the principles outlined by~\cite{gen-ai-paradox, ebt}, suggesting that discriminative, energy-based architectures exhibit greater robustness in OOD states than generative counterparts. We hypothesize that this robustness gap is further amplified in robotics, where OOD states naturally arise due to the stochastic and dynamic nature of real-world environments.

Moreover, this emergent behavior supports the theoretical arguments of EBT ~\cite{ebt} that energy-based verification and incremental energy minimization can yield models capable of native reasoning-like behavior. In repeated runs of the Tool Hang task, EBT-Policy consistently demonstrated this retry capability, recovering from perturbations that would terminate standard generative or behavior cloning policies.

To further analyze this phenomenon in Figure \ref{fig:dp_edp_compare}, we tracked EBT-Policy’s predicted energy trajectories across sampled robot rollouts showed in Figure \ref{fig:main}. The results revealed that the retry behavior closely corresponds to iterative energy minimization: the model initially enters high-energy states following a failed attempt, then progressively lowers its energy as it converges toward a stable, task-completing configuration. This alignment suggests that EBT-Policy’s emergent retry behavior is intrinsically tied to its uncertainty modeling and energy-based optimization. The findings further indicate emergent physical reasoning: EBT-Policy distinguishes successful from failed contact events and adapts its actions to re-establish effective interaction with the environment. Together, these results point to a compelling direction for future investigation into reasoning-driven robotic control.

In simulation, the maximum success rates achieved are somewhat lower than those reported in the original Diffusion Policy paper. This discrepancy is likely due to differences in hyperparameter configurations, while we matched the released real-world parameters as closely as possible, minor differences may have affected the simulated outcomes.

The main challenges we identified for EBT-Policy involve training stability and sampling under highly multimodal action distributions. While Diffusion Policy continues to outperform in stability and multimodal sampling, the new hyperparameter strategies and architectural refinements we introduce partially mitigate these issues. Future research should focus on improving optimization stability and mode coverage to fully leverage the benefits of energy-based implicit policies.
\section{Conclusion}
We presented EBT-Policy, an implicit policy that replaces iterative denoising with energy minimization over trajectories. By learning equilibrium dynamics, EBT-Policy delivers more stable inference, greater robustness to distribution shift, and substantially lower compute than diffusion- and flow-based baselines. On our simulated and real-world benchmarks, it consistently outperforms Diffusion Policy while using far fewer training epochs and as few as two inference iterations (up to 50$\times$ fewer than DP).

Beyond efficiency and accuracy, EBT-Policy exhibits emergent retry and adjustment after failure, guided by its scalar energy as an uncertainty signal. This behavior indicates interpretable physical reasoning: distinguishing successful from failed contacts and adapting actions to re-establish effective interaction. Together, these results lay a solid foundation for energy-based world models in manipulation, where a single energy function measures consistency across observations, language, actions, and dynamics.

It is excited to explore scaling EBT-Policy architectures, improved optimization schemes, and larger, more diverse datasets to further enhance performance and generalization. These directions cement EBT-Policy as a foundation for energy-based world models and advance unified, reasoning-capable robot policies under a single energy function.

{
    \small
    \bibliographystyle{ieeenat_fullname}
    \bibliography{main}
}


\end{document}